\documentclass[lettersize,journal]{IEEEtran}
\usepackage{booktabs} 
\usepackage{amsmath,amsfonts}
\usepackage{algorithmic}
\usepackage{algorithm}
\usepackage{array}
\usepackage[caption=false,font=normalsize,labelfont=sf,textfont=sf]{subfig}
\usepackage{textcomp}
\usepackage{stfloats}
\usepackage{url}
\usepackage{verbatim}
\usepackage{graphicx}
\usepackage{cite}
\begin{document}

\title{A 3M-Hybrid Model for the Restoration of Unique Giant Murals: A Case Study on the Murals of Yongle Palace}

\author{Jing Yang, Nur Intan Raihana Ruhaiyem, Chichun Zhou
\thanks{This work was supported by the Fundamental Research Grant Scheme FRGS/1/2021/ICT04/USM/02/1/Ministry of Higher Education. \emph{(Corresponding author: Nur Intan Raihana Ruhaiyem.)}

Jing Yang is with the School of Computer Sciences, Universiti Sains Malaysia, Gelugor, Penang 11800, Malaysia (yangjing666@student.usm.my). She is with the Shanxi Datong University, Datong, Shanxi, 037000, China.  

Nur Intan Raihana Ruhaiyem is with the School of Computer Sciences, Universiti Sains Malaysia, Gelugor, Penang 11800, Malaysia (intanraihana@usm.my).

Chichun Zhou is School of engineering, Dali university; Air-Space-Ground Integrated Intelligence and Big Data Application Engineering Research Center of Yunnan Provincial Department of Education, Yunnan, 671003, China.
}
\thanks{}}

\maketitle

\begin{abstract}
The Yongle Palace murals, as valuable cultural heritage, have suffered varying degrees of damage, making their restoration of significant importance. However, the giant size and unique data of Yongle Palace murals present challenges for existing deep-learning based restoration methods: 1) The distinctive style introduces domain bias in traditional transfer learning-based restoration methods, while the scarcity of mural data further limits the applicability of these methods. 2) Additionally, the giant size of these murals results in a wider range of defect types and sizes, necessitating models with greater adaptability. Consequently, there is a lack of focus on deep learning-based restoration methods for the unique giant murals of Yongle Palace. Here, a 3M-Hybrid model is proposed to address these challenges. Firstly, based on the characteristic that the mural data frequency is prominent in the distribution of low and high frequency features, high and low frequency features are separately abstracted for complementary learning.  Furthermore, we integrate a pre-trained Vision Transformer model (VIT) into the CNN module, allowing us to leverage the benefits of a large model while mitigating domain bias. Secondly, we mitigate seam and structural distortion issues resulting from the restoration of large defects by employing a multi-scale and multi-perspective strategy, including data segmentation and fusion. Experimental results demonstrate the efficacy of our proposed model. In regular-sized mural restoration, it improves SSIM and PSNR by 14.61\% and 4.73\%, respectively, compared to the best model among four representative CNN models. Additionally, it achieves favorable results in the final restoration of giant murals.
\end{abstract}


\section{Introduction}
\IEEEPARstart{T}{he} Yongle Palace murals represent outstanding artistic masterpieces in the history of painting. However, due to a lack of maintenance over time, these unique murals have developed numerous defects, making their restoration a pressing matter. Digital restoration methods have been proven to be more efficient and reversible compared to manual restoration techniques. In particular, image restoration techniques based on deep learning have achieved remarkable achievements. However, a comprehensive review of the literature on deep learning-based mural restoration reveals a predominant focus on Dunhuang murals\cite{wang2018dunhuang,yu2019end,Chen2021Mural,ciortan2021colour,xu2022deep} or other murals of regular size\cite{cao2020ancient,wang2021thanka}, with a lack of dedicated research specifically addressing the restoration of Yongle Palace murals and similar giant murals. When compared to other studies on mural inpainting, the restoration of unique giant murals in Yongle Palace faces two major challenges:\IEEEpubidadjcol

\begin{figure}[!t]
\centering
\includegraphics[width=3.5in]{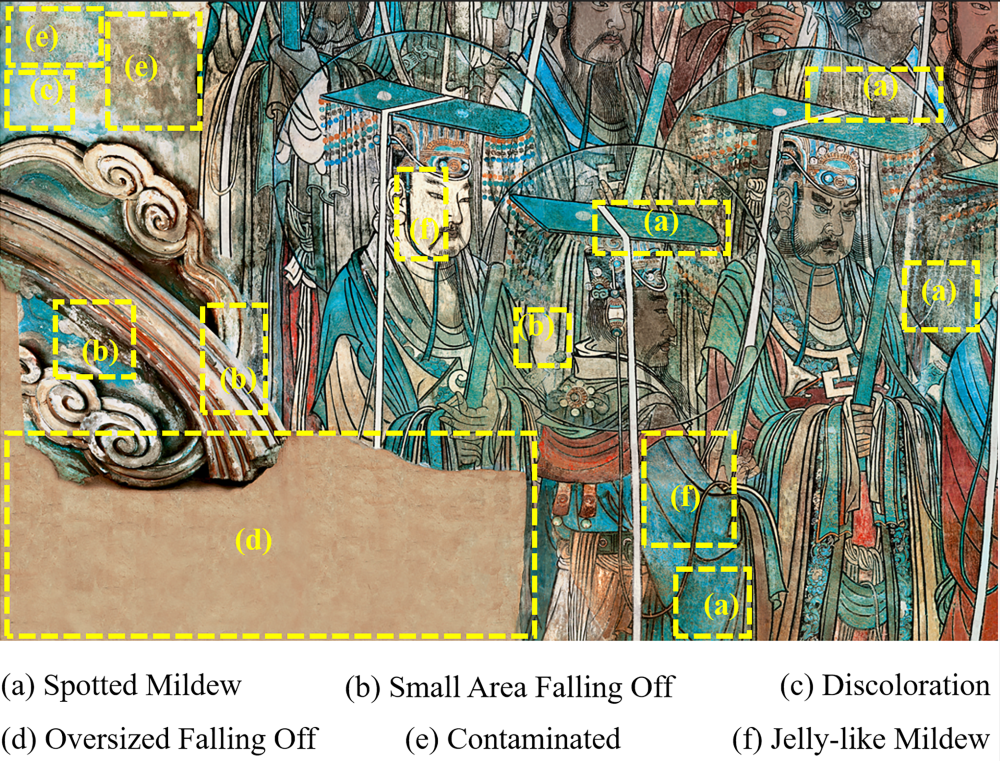}
\caption{Defects of Yongle Palace murals are varied in types and scales, and in this figure there are six of the types with different sizes indicated above, with the actual defects being more varied in presentation.}
\label{fig_1}
\end{figure}

\emph{Challenge 1:} The uniqueness of Yongle Palace murals lies in their scarcity and distinctive style. The limited availability of these murals poses challenges when training deep learning restoration models directly on a small dataset. Although transfer learning can help alleviate the issue of limited data, the distinctive style exhibited by Yongle Palace murals, characterized by abundant vivid large-area color blocks representing low-frequency features and sharply defined contours representing high-frequency features, significantly deviates from the common domain of images. These discrepancies hinder the effectiveness of transfer learning methods.

\emph{Challenge 2:} The giant size of murals results in a greater diversity of defect types and a wide range of defect sizes, particularly challenging oversized defects to be repaired. As illustrated in Fig.1, the immense size of the Yongle Palace giant murals increases the probability of encountering defects of different types and sizes. However, the model's proficiency in repairing defects of various types and sizes is limited. For instance, GMCNN\cite{wang2018image} excels at restoring rectangular blocks, while the PEN\cite{zeng2019learning} model is more adept at repairing randomly delineated lines. Thus, the objective of this study is to enhance the model's adaptability and capability to effectively restore a wide range of defects.

\begin{figure*}[!t]
\centering
{\includegraphics[width=7in]{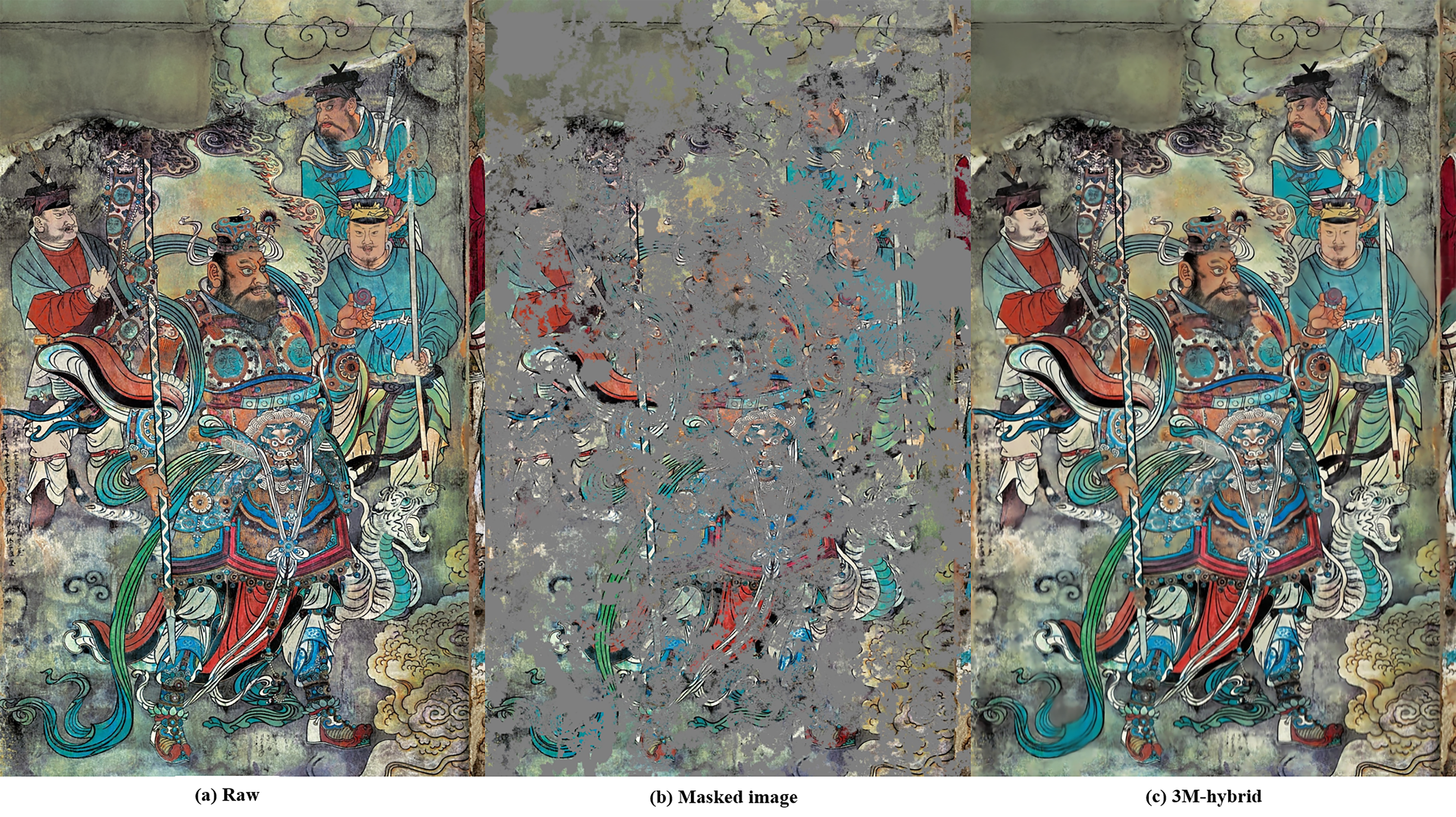}}%
\label{fig_first_case}
\hfil

\caption{(a) Raw refers to the original image of the giant mural. (b) Masked image refers to the damaged mural with added masks to indicate the areas of missing content. (c) 3M-Hybrid refers to the final restoration result of the giant mural achieved through the proposed 3M-Hybrid Model in this study.}
\label{fig_sim}
\end{figure*}

In order to accomplish the restoration of Yongle Palace murals without compromising the details, this study employs a method of segmenting the giant murals into regular-sized sections for repair and subsequently reassembling the restored sections back to their original dimensions. This implies that the success of the regular-sized mural restoration directly determines the outcome of the giant mural restoration as a whole. To enable the regular-sized mural restoration model to effectively address various types and scales of defects with a limited amount of unique image data, two aspects can be considered: optimizing the training data and refining the model structure. Regarding the optimization of training data, due to the scarcity of Yongle Palace murals, achieving desired results solely through conventional data augmentation techniques proves challenging. Therefore, it is crucial to conduct in-depth research on the data characteristics and adequately expose them during the data augmentation and training process to facilitate more effective feature learning by the model. In our investigation of the characteristics of Yongle Palace murals, we discovered that the images encompass abundant large-scale color blocks representing low-frequency information and well-defined contour lines representing high-frequency information. To achieve better restoration outcomes, this study employs separate networks dedicated to learning and extracting these high-frequency and low-frequency signals, thereby enhancing the feature learning and restoration capabilities in these specific frequency ranges. Furthermore, this frequency-based training approach enables the model to effectively address defects of different scales and types. In terms of the model structure, this study integrates Convolutional Neural Networks (CNN) with pre-trained Visual Transformers (VIT) to enhance the model's feature extraction capabilities. Furthermore, during the restoration of the giant mural, a basic cutting approach causes joint gaps and structural distortions when fixing oversized defects. To tackle this, we adopted a multi-perspective strategy to minimize gaps and used a multi-scale approach by combining cutting and downsizing methods. This ensures precise restoration and enhances extraction of the mural's overall structure while addressing multi-scale defects.

\subsubsection{Hybrid CNN-VIT network}

The hybrid CNN-VIT network integrates a pre-trained VIT feature extraction network into the CNN-based image restoration framework. The CNN network is selected to be the GLGAN\cite{iizuka2017globally} model which has the best performance in Yongle Palace mural restoration among the four models, namely, CE\cite{pathak2016context}, GLGAN\cite{iizuka2017globally}, GMCNN\cite{wang2018image}, and PEN\cite{zeng2019learning}. The VIT network is pre-trained on the ImageNet\cite{Imagenet} dataset. This integration not only preserves the powerful feature extraction capability of CNN on small datasets but also benefits from the enhanced effects and global information provided by the pre-trained VIT network.

\subsubsection{Multi-frequency strategy}

This study focuses on the restoration of Yongle Palace murals by extracting and restoring the high-frequency and low-frequency information using a hybrid CNN-VIT network. The network is designed to independently learn and restore the high-frequency, low-frequency, and full-frequency images. The restoration process involves combining the results obtained from each frequency domain and feeding them as inputs to a second-stage network, ultimately leading to the final restoration outcome.

\subsubsection{Multi-perspective strategy}

The multi-perspective strategy overcomes the seam artifact issue caused by the excessive size of the damaged giant murals by cutting them into 16 different versions of regular-sized images from various perspectives and then reassembling them after restoration. This approach compensates for the missing information between images through different cuttings and effectively smoothens the transition seams using an averaging strategy. Moreover, by adopting different cutting perspectives, more favorable restoration segments can be identified, thereby optimizing the overall restoration outcome.

\subsubsection{Multi-scale strategy}

The multi-scale strategy effectively addresses the issue of overall and structural information loss when restoring oversized defects. By adopting a multi-scale approach, the murals are downsized to 4/5 and 3/5 of the original size, in addition to the original size, providing three distinct mural sizes. The mural at the original size is responsible for detail restoration, while the downsized murals offer structural information at different scales. By combining the results from the three scales, a more reliable final restoration outcome can be achieved. Simultaneously, this multiple scales can effectively mitigate the multi-scale defect repair.

Based on the above, a series of experiments were conducted in this study, and the result of the restoration for the unique giant mural in Yongle Palace is shown in Fig.2. The experimental results fully demonstrate the effectiveness of the proposed approach. Finally, the contributions of this research are as follows:

Contribution 1:  This study investigates the application of deep learning in the restoration of giant murals, with a particular focus on utilizing deep learning techniques for restoring Yongle Palace murals. This research represents the first attempt to explore deep learning-based restoration methods for such large-scale artworks.

Contribution 2: In improving the regular-sized image restoration models, this study integrates comprehensive improvements from both data and structure perspectives. This provides new insights for future research on the restoration of unique small datasets.

\section{Related Work}
\subsection{Image Inpainting Methods}

Mural restoration is a specific application of image restoration. Digital image restoration models can generally be classified into two categories: traditional methods and deep learning-based methods. Extensive research consistently demonstrates that, when addressing complex problems, deep learning models outperform traditional models. Therefore, this study specifically focuses on deep learning-based image restoration models. Deep learning-based image restoration algorithms can be categorized into several types, including: 1) encoder-decoder models; 2) U-network models; 3) generative adversarial network (GAN) models; 4) transformer models; and 5) denoising diffusion models.

\subsubsection{Encoder-Decoder class} The Context Encoder (CE) model\cite{pathak2016context}, as a preliminary exploration of the Encoder-Decoder framework, is a type of GAN model that emphasizes intra-image inpainting based on contextual information. Yang et al.\cite{yang2017high} replaced convolutional layers with residual blocks in the CE model. The Edge-Aware Context Encoder model\cite{liao2018edge} builds upon the Context Encoder model by incorporating edge restoration techniques.

\subsubsection{GAN class} The GLGAN model\cite{iizuka2017globally} incorporates global and local discriminators to enhance image restoration effectiveness. The StackGAN\cite{zhang2017stackgan} proposes a two-stage generative adversarial network that progressively improves the restoration results from coarse to fine. The GMCNN model\cite{wang2018image} utilizes convolutional kernels of different scales to capture image features with various receptive fields. The EdgeConnect\cite{nazeri2019edgeconnect} model incorporates edge information into image restoration. The WaveFill\cite{yu2021wavefill} model introduces the concept of wavelet transform, performing image restoration in multiple frequency bands and employing frequency domain attention normalization to align and fuse multi-frequency features.

\subsubsection{UNet class} The Shift-Net model\cite{ yan2018shift} integrates a shift-join layer into the U-Net architecture, facilitating the restoration of complex structures and intricate textures within arbitrarily shaped defective regions. The DFNet model\cite{ hong2019deep} incorporates fusion modules within the last five decoding layers to ensure a seamless transition between restored and intact areas. The PEN model\cite{ zeng2019learning} utilizes a pyramidal context encoding network to capture contextual semantics from high-resolution inputs and decode the learned semantic features for inpainting purposes. The HiFill model\cite{yi2020contextual} generates high-frequency residuals through weighted aggregation of residuals from contextual samples, treating them as missing content, while the attention module calculates attention scores and performs attention transfer within the U-Net structure. This design significantly enhances image restoration quality across multiple scales, including up to 1024$\times$1024 pixels in size.

\subsubsection{Transformer class} Zhou et al.\cite{zhou2021transfill} pioneered the introduction of a transformer-based inpainting model, tailored to address the challenges of repairing complex scene images. Following this, Wang et al.\cite{wan2021high} proposed a two-stage inpainting method that combines the reconstruction of appearance priors using Transformers and the supplementation of textures using CNNs. The utilization of Masked Autoencoders\cite{he2022masked} brought forth the integration of self-supervised learning in the field of computer vision. Additionally, Dong et al.\cite{dong2022incremental} developed an incremental transformer structure inpainting network that incorporates masked position encoding to improve the model's ability to generalize across various mask types.

\subsubsection{Denoising diffusion} Denoising diffusion inpainting models currently depend on large-scale datasets and high-performance computing resources for effective operation. Prominent algorithms within this category encompass the Denoising Diffusion Inpainting Model (DDIM)\cite{song2020denoising} and the Denoising Diffusion Probabilistic Model (DDPM)\cite{lugmayr2022repaint}, both of which have garnered considerable attention and widespread adoption in the field.

\subsection{Mural Inpainting Methods}

There have been limited studies on deep learning-based mural restoration. Wang et al.\cite{wang2018dunhuang} proposed a Circle GAN model to address minor defects in Dunhuang murals. Yu et al.\cite{yu2019end} employed a U-Net with partial convolutional layers to reconstruct Dunhuang murals. J. Cao et al.\cite{cao2020ancient} presented a GLGAN model with FCN for addressing small localized defects in Wutaishan murals. Chen Yong et al.\cite{Chen2021Mural} proposed a multi-scale kernel U-Net model based on Dunhuang murals, which first restores the structure and then the details. N. Wang et al.\cite{wang2021thanka} introduced a U-Net model with multi-scale kernels for restoring Thangka murals. Ciortan et al.\cite{ ciortan2021colour} developed a two-stage GAN model based on Dunhuang murals, where the first stage focuses on edge restoration, the second stage handles color restoration, and the final step involves color adjustment.W Xu and Y Fu\cite{xu2022deep} proposed a colour restoration method for Dunhuang images based on DenseNet algorithm.

\subsection{Research Gap}

In terms of image restoration models, according to the research conducted by Luo\cite{luo2022review} and Li\cite{LI2021review}, as well as our limited understanding, the maximum image size that image restoration models can handle is 2048×1024 pixels. However, the digitized dimensions of the Yongle Palace murals far exceed this area. Therefore, existing models cannot directly restore large-scale murals. Furthermore, the image datasets used by current image restoration models primarily consist of ImageNet\cite{Imagenet}, Places2\cite{Places2}, and CelebA-HQ\cite{CelebA-HQ}, which contain 14197122, 1000000, and 202599 images, respectively, far exceeding the quantity of Yongle Palace murals. Among the five discussed models, the transformer model and the denoising diffusion model have emerged in recent years and achieved remarkable results in large-scale tasks. However, in specific domain applications, these large-scale models often face the limitation of scarce data, especially in mural restoration tasks. Mural samples are not only limited in quantity but also possess unique styles, making it challenging to directly transfer them into such large-scale restoration models. Therefore, this research aims to address the problem of how to adapt the model to small-scale datasets like CNN models while leveraging the feature extraction capabilities of the best-performing models, such as transformers.

In terms of mural restoration models, the focus primarily lies on CNN models, specifically GAN and U-Net image restoration models. While these models exhibit impressive performance in addressing image restoration challenges on small datasets, they are inevitably constrained by the inherent characteristics of CNNs and their limited ability to capture global features. Furthermore, the majority of existing deep learning-based research on mural restoration has centered around Dunhuang murals, with limited attention given to the restoration of Yongle Palace murals. Additionally, the current studies have predominantly concentrated on small-scale murals or specific sections, lacking dedicated research on the comprehensive restoration of large-scale murals. In reality, large-scale murals present significantly more complex challenges that warrant particular attention.

\section{METHODOLOGY}

\begin{figure*}[!t]
\centering
{\includegraphics[width=7in]{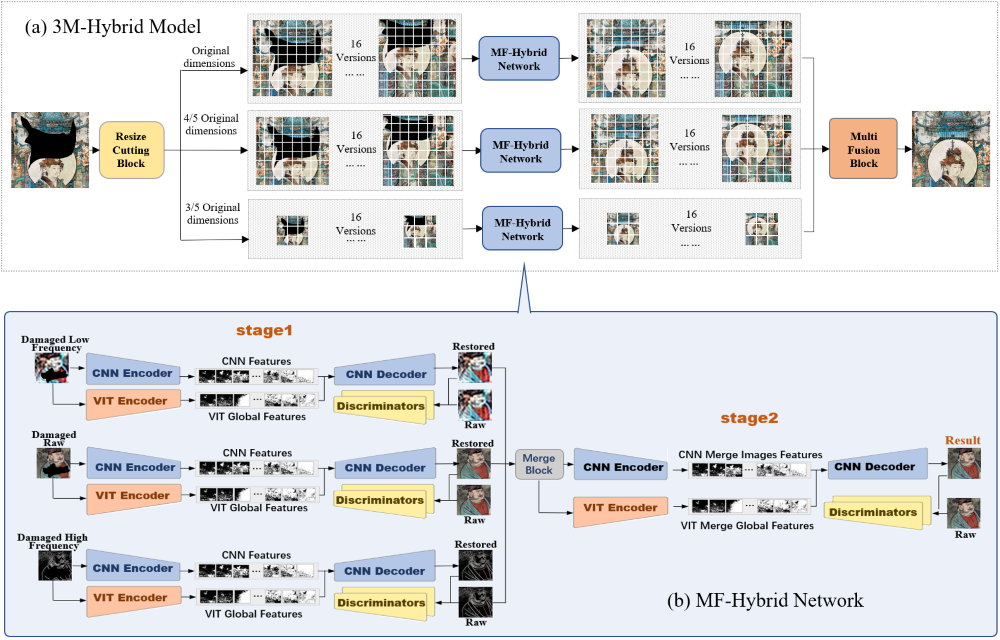}%
\label{fig_first_case}}
\hfil
\caption{(a) 3M-Hybrid Model: Firstly, the Resize Cutting Block downsizes the giant mural using a multi-scale strategy into two sizes: 4/5 and 3/5 of the original size. Together with the original size, these become three different scales of large, medium, and small murals. Then, the murals are further cut into 16 versions of regular-sized murals, each sized 256$\times$256 pixels, using a multi-perspective strategy. Secondly, the small murals are restored using the trained MF-Hybrid Network to obtain the restored small murals. Finally, the Multi Fusion Block assembles the restored small images using a multi-perspective and multi-scale strategy to obtain the final result. (b) MF-Hybrid Network: The MF-Hybrid Network, based on regular-sized mural restoration, is divided into two stages. In the first stage, the images in the mural dataset are processed separately for low and high frequencies. The low-frequency image, high-frequency image, and original image of the mural are obtained as inputs to the hybrid CNN-VIT network for restoration. The backbone of the hybrid CNN-VIT network is GLGAN, which has proven to be the most effective in experiments for the restoration of the Yongle Palace murals. A pre-trained VIT enhancement module is added to the feature extraction process to enhance the model's feature extraction capability. In the second stage, restoration results from three image restoration networks, focusing on different frequency bands (i.e., high frequency, low frequency, and full frequency), are fed into a merge block. The merge block reduces the nine channels to three using a learnable single-layer convolutional network, and the results are then fed into a hybrid CNN-VIT network similar to the first stage. This way, the image restoration results from the three deep learning networks are fused and re-learned to obtain the final mural restoration results.}
\label{fig_sim}
\end{figure*}

To address the restoration challenges of unique giant murals, this study proposed a 3M-Hybrid Model (Multi-frequency, Multi-viewpoint, Multi-scale CNN and VIT Hybrid Model), as illustrated in Fig.3. The restoration process involves three main steps.

Firstly, split the giant mural into regular sized sections. Following the multi-scale strategy, the giant damaged mural is resized into three scales: the original size, 4/5 of the original size, and 3/5 of the original size. Subsequently, employing the multi-perspective strategy, these three differently scaled murals are further segmented into 16 regular-sized mural images of 256$\times$256 pixels.

Secondly, restoration of regular-sized murals. Utilizing the MF-Hybrid network composed of a multi-frequency strategy and a hybrid CNN-VIT network restores the regular-sized murals.

Thirdly, reassembling restored regular-sized murals into giant murals. Employing the multi-perspective strategy, the 16 repaired versions of each different scale is averaged individually. Then, the averaged results from the three scales are combined with appropriate weights to form the final restoration output of the giant mural.

The subsequent sections provide a detailed explanation of the hybrid CNN-VIT network and the multi-frequency, multi-perspective, and multi-scale strategies described above.

\subsection{MF-Hybrid Network}

The restoration of giant murals ultimately relies on the performance of regular-sized mural restoration models. However, the murals in the Yongle Palace are scarce in quantity, distinctive in style, and exhibit various types of defects at multiple scales, posing challenges for regular-sized mural restoration models. To improve the restoration models, two approaches can be employed. The first approach is to increase the quantity and quality to expand the available information from data. The second approach is to optimise the model structure to enhance the ability of features extraction. Therefore, for the restoration of regular-sized murals, we propose a Multi-Frequency Hybrid Network (MF-Hybrid Network), as shown in Figure 3(b), to enhance restoration capabilities from both data and structural perspectives.

\subsubsection{Hybrid CNN-VIT Network}   \quad

As widely known, CNN extracts local to global information from images through a series of overlapping convolutional layers, gradually expanding the receptive field until it covers the entire image. CNN has a great advantage in image feature extraction, allowing features to be extracted quickly on a small amount of data and exhibiting an inherent bias towards certain image properties, such as translation invariance. However, advanced visual semantic information often requires understanding the relationships between elements, such as object formation and spatial positioning within a scene. Transformers, on the other hand, have access to global information from the outset, rather than starting from local information. Despite the training challenges, Transformers have the capability to capture longer-range dependencies. Therefore, in this study, we employ a combination of CNN as the backbone and Transformer as an enhanced feature extractor to address mural restoration tasks.

Due to the data limitations of the Yongle Palace murals dataset in this study, which prevent the direct utilization of a single Vision Transformer (VIT)-based image restoration model for mural restoration, the Transformer structure can only supplement the CNN model. In this research, through training on the Yongle Palace murals, the model parameters are adjusted, and the attention mechanism is utilized to augment the CNN with global features. A pre-trained VIT model based on ImageNet\cite{Imagenet} is used to enhance the model's restoration capabilities. This model incorporates a 12-layer encoder from the pre-trained VIT model to transform the extracted one-dimensional features into two-dimensional feature representations. Finally, the features extracted by the VIT model are concatenated with the features extracted by the original CNN network and passed to the decoder for further processing and interpretation.

Furthermore, in order to maximize the performance of the model, the selection of the CNN model, which serves as the backbone of the restoration model, is crucial. To determine which model is most suitable for the restoration of Yongle Palace murals, we conducted a summary of the existing mural restoration algorithms' structures and found that these restoration models can be categorized into UNet-based models and GAN-based models, enhanced with multi-scale convolutional kernels, global and local discriminators, and attention mechanisms. Considering that most mural restoration studies have not provided publicly available code, we combined the research of Li Yuelong \cite{LI2021review} and evaluated several models with similar structures that showed good restoration results on small datasets. Based on the experimental results (see Appendix for details), we selected the GLGAN\cite{iizuka2017globally} model from GLGAN\cite{iizuka2017globally}, CE\cite{pathak2016context}, GMCNN\cite{wang2018image}, and PEN\cite{zeng2019learning} models as the CNN network structure for the MF-Hybrid Network, which is the most suitable for the restoration of Yongle Palace murals.

\subsubsection{Multi-frequency strategy} \quad 

In terms of data, the girant murals of the Yongle Palace can be divided into about 5000 regular-sized images with dimensions of 256$\times$256 pixels. In comparison to image databases encompassing millions of samples, the dataset of Yongle Palace murals is relatively limited in size, thereby presenting challenges for prevailing large-scale models engineered to deliver remarkable outcomes. Furthermore, the unique artistic style of Yongle Palace murals, combined with their diverse range of defects encompassing various types and scales, makes it challenging to directly apply pre-trained  models to achieve the desired restorative effects. Consequently, the proposed model in this study strives to effectively harness the limited data available from the Yongle Palace murals, proficiently extract their distinct features, and ultimately achieve restoration results of superior quality.

The murals of Yongle Palace exhibit a plethora of vibrant color blocks and well-defined edge information. These color blocks represent long-distance low-frequency features, while the edges capture short-distance high-frequency characteristics. As a result, the frequency distribution of mural images contains a significant amount of information in both high and low frequencies. Therefore, it is imperative for the model to enhance its learning capability for both high-frequency and low-frequency features. Inspired by the Edgeconnect\cite{nazeri2019edgeconnect} model and edge-aware context encoding\cite{liao2018edge} model, we propose the inclusion of a high-frequency and low-frequency restoration network to bolster the model's learning capacity for low-frequency features.

Furthermore, according to the research conducted by Robert Geirhos et al.\cite{geirhos2018imagenet}, CNNs tend to prioritize texture over shape feature extraction. Training models based on shape feature extraction can enhance the model's ability to extract more robust human-like features, thereby improving its restoration capabilities. The application of high-frequency and low-frequency processing to images is also a fundamental texture removal technique mentioned in previous studies. This approach indirectly strengthens the model's ability to extract shape features, resulting in superior restoration outcomes.

The multi-frequency strategy effectively adapts to the frequency characteristics of different images, thereby facilitating the learning of diverse data features. By decomposing the complex task of extracting full-band features into simpler sub-band feature extraction tasks, the proposed model achieves reinforced learning of relevant features, leading to superior restoration outcomes. Moreover, the inclusion of high-frequency and low-frequency texture removal features further enhances the model's ability to capture shape features, thereby improving the overall image restoration capability. Lastly, the multi-frequency approach implies multi-scale, allowing the restoration of features at different scales, which is particularly advantageous for repairing multi-scale and multi-type defects.

\subsection{Assembly of a giant mural}

Previous research on mural restoration based on deep learning has primarily focused on regular-sized murals or partial restoration of giant murals. However, there is a notable research gap when it comes to the restoration of complete giant murals. The restoration of giant murals presents significant challenges compared to regular-sized ones, as the size of the giant mural to be restored exceeds the capacity of the existing model, and oversized defects also pose difficulties in finding suitable solutions. Cropping and downsizing are two methods of converting the restoration of giant murals into regular-sized murals, but both have drawbacks.

Cropping can basically fixes small imperfections on giant murals, which preserves fine details but may result in the loss of global structural features and produce seams at the boundaries of the cropped sections when confronted with oversized defects. On the other hand, downsizing the mural is more effective in preserving the overall structural features but may lead to a significant loss of fine details, which can compromise the restoration quality required for mural detail restoration. Therefore, in this research, we employ the cropping method as the primary approach to retain the details of the mural while employing a multi-perspective strategy to address the issue of seams. Additionally, a multi-scale strategy using the downsizing method is employed as an auxiliary approach to enhance the restoration performance of the model, particularly for defects of extremely large size. The specifics of the multi-perspective and multi-scale strategies are as follows:

\subsubsection{Multi-perspectives strategy}	  \quad

\begin{figure}[!t]
\centering
\includegraphics[width=3.6in]{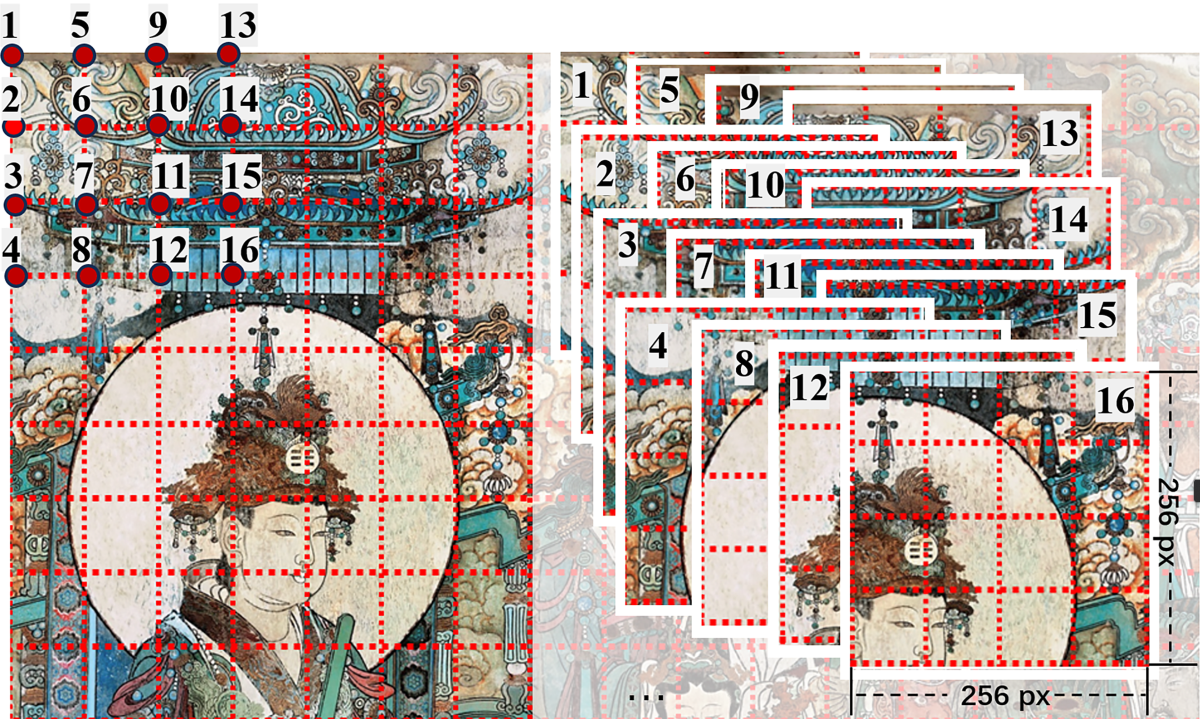}
\caption{Crop the giant mural image into smaller images of 256$\times$256 pixels in size in order, following the marked positions of points 1, 2 and up to 16 in the diagram respectively. This will give us 16 versions of the small mural set after cropping.}
\label{fig_2}
\end{figure}

The multi-perspective strategy is primarily employed to address the seam restoration issue. When a giant mural is divided into smaller regular-sized murals, continuous defects may also be fragmented into multiple parts. The inter-image correlation information between these smaller images is lost, resulting in a lack of constraints and guidance during defect restoration. Inconsistent feature allocation for continuous defects across different smaller images leads to noticeable seams and discontinuities at the boundaries.

To better capture the internal relationships within the images and effectively utilize the structural connections between each image, this study proposes dividing the giant mural into 16 versions of smaller images from different viewpoints, as shown in Fig.4. The 16 different segmentation patterns allow each defect to acquire diverse contextual information, indirectly providing the inter-image correlation and positional information for the model. When faced with the same defect, even if some viewpoints yield unsatisfactory restoration results, the averaging operation of the 16 versions can generate smoothly transitioning restoration outcomes, eliminating seam artifacts.

Furthermore, by utilizing these 16 different viewpoints, excessively large defects can be transformed into smaller defects with varying sizes and viewpoints. Therefore, even if certain regions in the segmented mural contain defects that are too large to be individually restored, the results obtained from the smaller images segmented from different viewpoints can be utilized for restoration purposes.

\subsubsection{Multi-scale strategy}	  \quad
 
In the context of restoring giant murals, researchers are confronted with the challenge of addressing oversized defects within the giant murals. When the study segments the giant murals into smaller images for restoration purposes, it becomes inevitable that each small image lacks access to long-distance information beyond its own size after the segmentation process. Moreover, certain small images may contain defects that exceed the limited information present within the image itself, thereby impeding the model's ability to achieve high-quality restoration and resulting in weak or infeasible results.

To alleviate these challenges and address the issue of oversized defects, this study proposes a multi-scale strategy. Within this strategy, the image is resized into three different scales: the original size, 4/5 of the original size, and 3/5 of the original size. Scaling the image to smaller sizes increases the coverage of the cropped small image, allowing for a wider field of view, enhanced acquisition of global information, and better comprehension of long-distance features, thereby improving the overall structure of the restored full image.

However, it is important to note that scaling the image results in the loss of detailed information, which constrains the delicacy of the restoration outcomes. To strike a balance between the delicacy of image restoration and the coherence of its structure, this study combines the restoration results obtained from giant murals at three scales (original, 4/5, and 3/5) using specific weights. For Fig.2, the weights are set as 0.8, 0.1, and 0.1, while for Fig.7, the weights are set as 0.7, 0.2, and 0.1. These weight settings have been derived from the results of many trials and may vary depending on the specific mask being used. Generally, when dealing with larger and more oversized defects in the masks, a higher proportion of small scales should be assigned to achieve better restoration results.

\section{EXPERIMENTAL RESULTS AND ANALYSIS}

\subsection{Experimental settings} 
The experimental GPU is RTX 3090 (24GB), based on CUDA Version 11.6, python 3.8.10, and Pytorch version is 1.11.0+cu113.

The training process of the MF-Hybrid model consists of two phases. During the training phase, the initial learning rate is set to 0.0001. The training process continues until there is no further decrease in the loss, at which point the learning rate is reduced by multiplying it by 0.1 to facilitate further training. This process continues until the learning rate reaches 0.00000001, at which stage the training is finally terminated.

\subsection{Experiment dataset and evaluate metrics}

\subsubsection{Experimental mural data} The Yongle Palace mural can be divided into 10 giant murals with the following sizes: 2560$\times$4096 pixels, 14848$\times$4608 pixels, 12800$\times$4608 pixels, 5120$\times$4608 pixels, 4096$\times$4608 pixels, 4096$\times$4608 pixels, 4096$\times$4352 pixels, 11776$\times$4608 pixels, 15360$\times$4608 pixels, and 2560$\times$4096 pixels. In this study, the first giant mural was selected as the test set, while the remaining murals were used as the training set. Before training, all the giant mural data were scaled down to 4/5 and 3/5 of their original size to implement the multi-scale completion strategy. Subsequently, each image, at the three different scales, was divided into 16 versions of small mural images with a size of 256$\times$256 pixels, as part of the multi-perspective integration strategy. Following this, the model performed low-frequency and high-frequency edge extraction on these small mural images. The low-frequency image of the mural was obtained through Fourier transform, while the high-frequency image was obtained using a Sobel filter.

\begin{figure}[!t]
\centering
\includegraphics[width=3.5in]{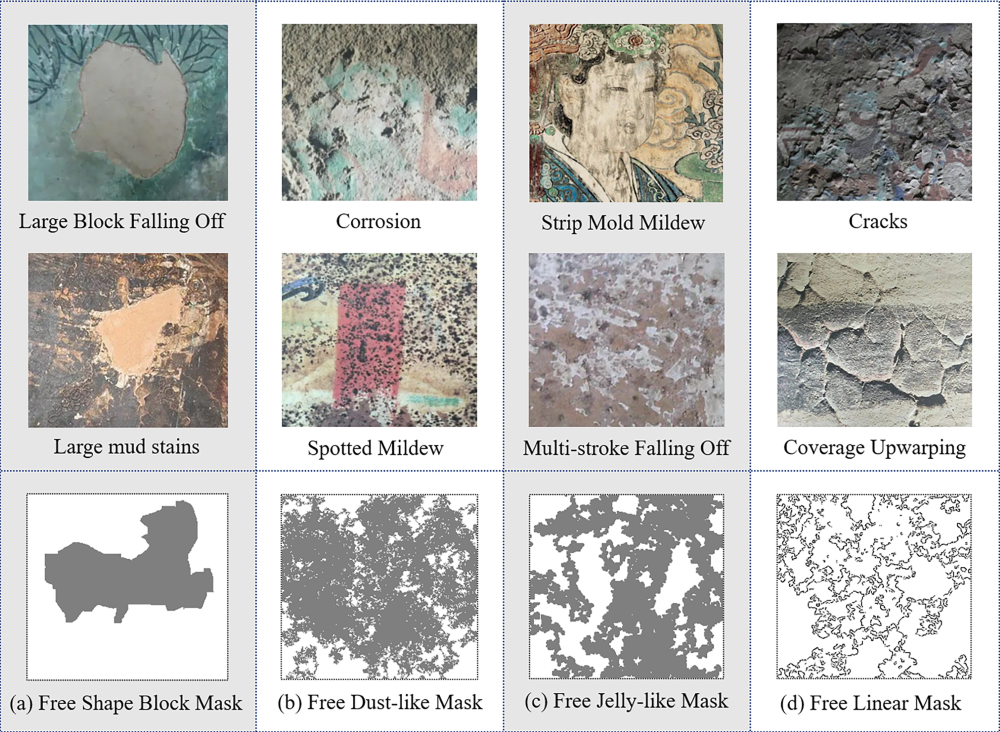}
\caption{(a) The free shape block mask is designed to simulate large areas of random-shaped defects on the mural, such as falling-off blocks and mud stains. These defects occur in irregular shapes and sizes, making them challenging to restore. (b) The free dust-like mask, proposed by T. Yu for Dunhuang Mural inpainting \cite{yu2019end}, is generated by randomly selecting a starting point and setting the pixel value to 1. This mask simulates corrosion and spotted mildew, which are common forms of deterioration on mural surfaces. (c) The free jelly-like mask, also proposed by T. Yu for Dunhuang murals \cite{yu2019end}, expands irregular defective regions by removing small noises and applying an erosion function. This mask is used to simulate strip mold mildew and multiple stroke falling-off effects. (d) The free linear mask, proposed by Ciortan for Dunhuang Mural inpainting \cite{ciortan2021colour}, includes two types based on the dust-like mask. The first type is generated by skeletonization of the dust-like mask, while the other type is generated through medial axis transform and dilation of the dust-like mask. These masks simulate different linear defects, such as coverage upwarping and cracks.}
\label{fig_2}
\end{figure}

\subsubsection{mask data} Based on the studies of murals in Yongle Palace, the main types of damage we need to address are falling off, corrosion, mildew, and cracks. However, to make our research more comprehensive, we also included common damages found in other murals, such as coverage upwarping and mud stains. The classification of mural damage in the field of archaeology and cultural relic restoration is primarily based on the causes and methods of repair, rather than the specific shape and size characteristics. In the context of computer vision-based image restoration, we are primarily concerned with the shape and size of defects. Therefore, we employ four types of masks to simulate common mural damages, as shown in Fig.5. These include the free shape block mask, the free dust-like mask \cite{yu2019end}, the free jelly-like mask \cite{yu2019end}, and the free linear mask \cite{ciortan2021colour}. The experiments described below were conducted using these four types of masks at different scales.

\subsubsection{Evaluate metrics} In this research, four evaluation metrics have been employed to assess the performance of the models. These metrics include Mean Absolute Error (MAE), which quantifies image error; Mean Squared Error (MSE), which measures image similarity; Peak Signal to Noise Ratio (PSNR), which evaluates image distortion; and Structural Similarity Index (SSIM), which measures image similarity based on factors such as luminance, contrast, and structural attributes.

\subsection{Experimental results}

\begin{figure*}[!t]
\centering
{\includegraphics[width=7in]{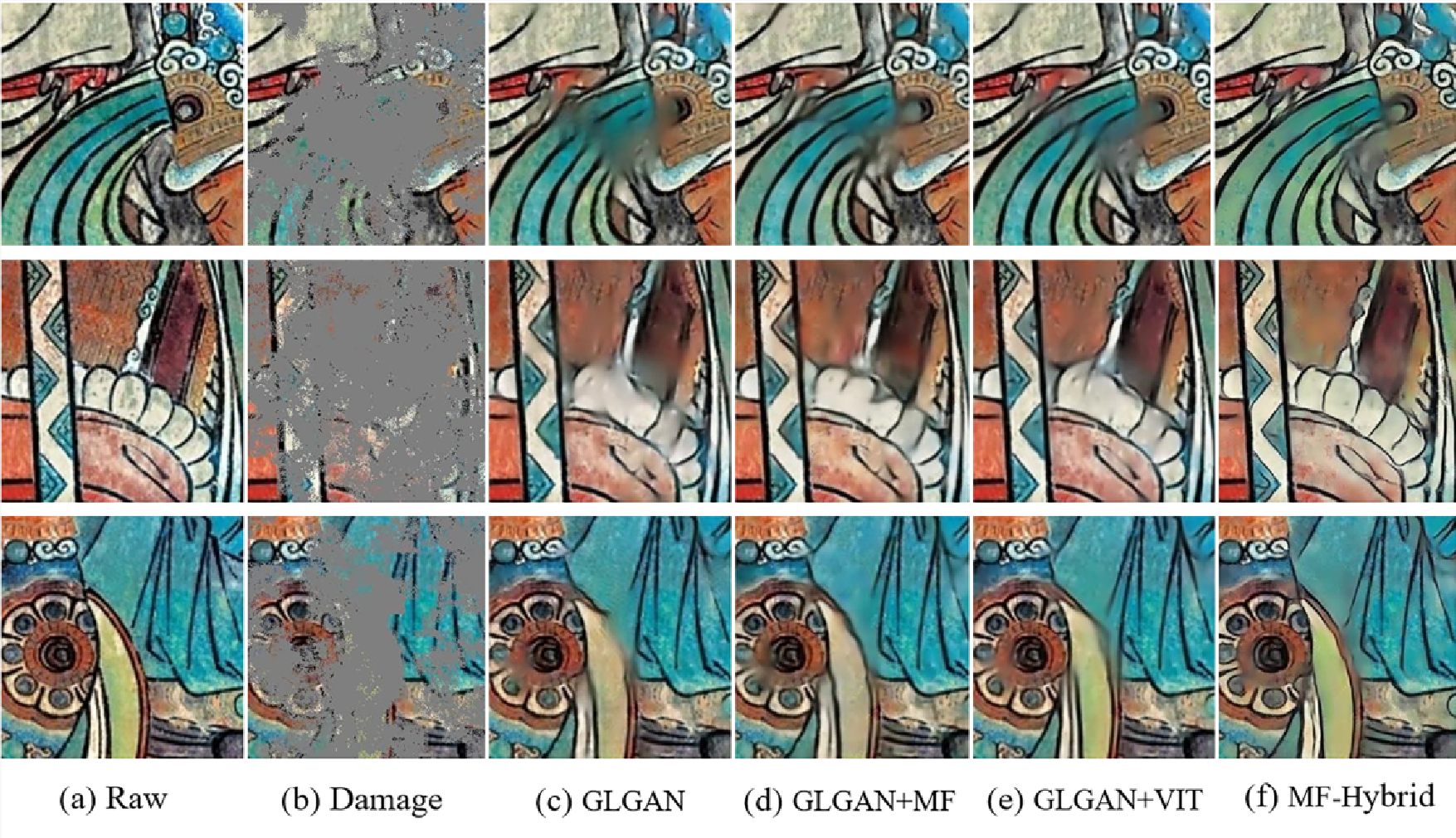}%
\label{fig_first_case}}
\hfil
\caption{(a)Raw is the raw mural images. (b) Damage refers to the masked images. (c) GLGAN is the result restored by GLGAN model. (d) GLGAN+MF refers to the results restored by the GLGAN model with a multi-frequency strategy. (e) GLGAN+VIT is the result of restoration of a hybrid CNN-VIT network based on GLGAN with the addition of a VIT feature extraction network. (f) MF-Hybrid refers to the results restored by the proposed MF-Hybrid network which is the network based on GLGAN with the multi-frequency strategy and a VIT feature extraction network.}
\label{fig_sim}
\end{figure*}

The presented 3M-Hybrid model, an extension of the regular-sized mural restoration model MF-Hybrid, seamlessly integrates multi-perspective and multi-scale strategies to accomplish the restoration of giant murals. The restoration outcomes achieved by the 3M-Hybrid model are illustrated in Fig.2. From the visual representations, it is evident that the proposed model exhibits commendable restoration results for free dust-like, free jelly-like, and free linear masks. Moreover, the restoration outcomes for the free shape block mask demonstrate preserved structural integrity and plausible texture. Based on these findings, we can confidently conclude that the 3M-Hybrid model proves to be a viable approach for restoring these unique and giant murals.

\subsection{Ablation experiments}

To demonstrate the effectiveness of the hybrid CNN-VIT network, multi-frequency strategy, multi-perspective strategy, and multi-scale strategy used in the proposed 3M-Hybrid Model, a series of ablation experiments were conducted in this study.

\subsubsection{The regular-sized mural ablation experimental of MF-Hybrid network}

\begin{table}[]
    \caption{Comparison of the results of GLGAN and MF-Hybrid Network\label{tab:table1}}
    
    \begin{tabular}{ccccc}
    \toprule[2pt]
    \makebox[0.07\textwidth][c]{INDEX} & \makebox[0.07\textwidth][c]{MODEL} & \makebox[0.07\textwidth][c]{LAYER 1} & \makebox[0.07\textwidth][c]{LAYER 2} & \makebox[0.07\textwidth][c]{LAYER 3} \\ \hline
    
    \hline
              
    ~ & GLGAN & 17.221491 & 15.375962 & 14.271284 \\
    PSNR & GLGAN+VIT	   & 18.146103 & 15.875593 & 15.02639 \\
    ~   & MF-Hybrid    & \pmb{18.841078} & \pmb{17.806002} & \pmb{17.068996} \\ \hline
    ~ & GLGAN        & 0.697161  & 0.612244  & 0.571007  \\     
    SSIM & GLGAN+VIT    & \pmb{0.727926}  & 0.632828  & 0.612619  \\
    ~   & MF-Hybrid    & 0.691353  & \pmb{0.642969}  & \pmb{0.63498}   \\ \hline
    ~   & GLGAN        & 0.078318  & 0.102102  & 0.117713  \\
    MAE  & GLGAN+VIT    & \pmb{0.069497}  & 0.095515  & 0.105342  \\
    ~   & MF-Hybrid    & 0.070418  & \pmb{0.081945}  & \pmb{0.088854}  \\ \hline
    ~   & GLGAN        & 0.024364  & 0.036155  & 0.044133  \\  
    MSE  & GLGAN+VIT    & 0.020107  & 0.03240  & 0.036848   \\
    ~   & MF-Hybrid    & \pmb{0.015803}  & \pmb{0.0199}    & \pmb{0.022535}  \\ \hline
    \bottomrule[2pt]
    \end{tabular}
\end{table}

\begin{figure*}[!t]
\centering
{\includegraphics[width=7in]{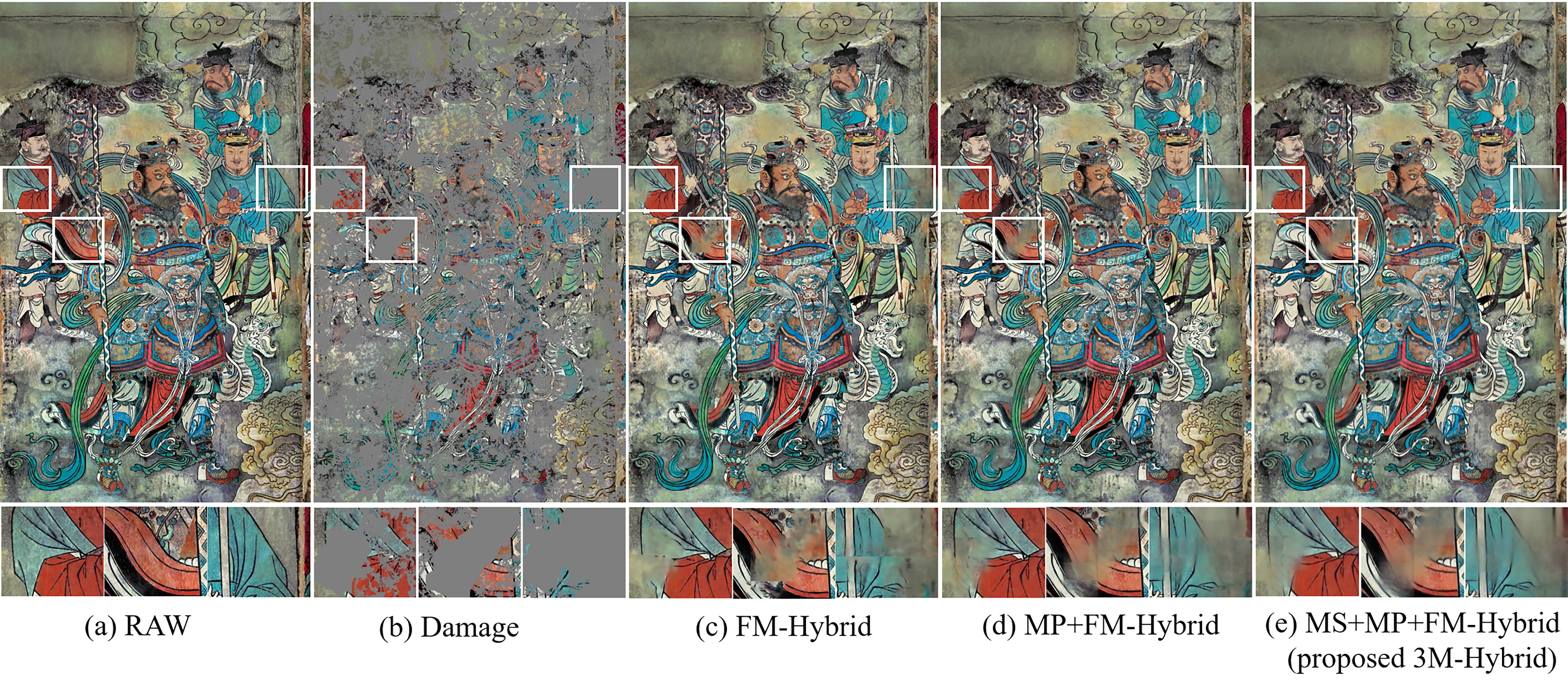}%
\label{fig_first_case}}
\hfil
\caption{(a) RAW is the original mural. (b) Damage is the masked image. (c) "FM-Hybrid" is the result restored by FM-Hybrid with regular-sized splicing. (d) "MP+FM-Hybrid" is the result restored by FM-Hybrid with multi-perspective strategy splicing. (e) "MS+MP+FM-Hybrid (proposed 3M-Hybrid)" is the result restored by the proposed 3M-Hybrid Model which means repaired by FM-Hybrid and spliced by multi-perspective and multi-scale strategies. }
\label{fig_sim}
\end{figure*}

The results of the proposed MF-Hybrid network for regular-sized mural inpainting are shown in Fig.6. The MF-Hybrid Network includes hybrid CNN-VIT networks and multi-frequency strategies. It is evident from the diagram that both of these improvements have enhanced the restoration of regular-sized murals, and when used together, they yield better results than the previous three.

In addition to the evident visual improvements of the MF-Hybrid network over the GLGAN network, its advancements can also be quantitatively observed from the data. This study employs four evaluation metrics, namely PSNR, SSIM, MAE, and MSE, to compare the restoration results of GLGAN and the proposed MF-Hybrid network for small-sized murals. The comparison results are presented in the TABLE \MakeUppercase{\romannumeral 1}.

TABLE \MakeUppercase{\romannumeral 1} includes the LAYER 1, which represents the restoration of small images cropped from the original-sized giant murals, the LAYER 2 denotes the restoration of small images scaled down to 4/5 of the original size, and the LAYER 3 signifies the restoration of small images scaled down to 3/5 of the original size. After adopting the VIT enhancement strategy, GLGAN exhibits significant improvements across all metrics. The MF-Hybrid network, building upon the improvements of the GLGAN network, not only incorporates the VIT enhancement strategy but also leverages the multi-frequency complementation strategy. Through comprehensive evaluation using these four metrics, it becomes evident that the MF-Hybrid network demonstrates substantial improvements in restoring LAYER 1 and 2. While both the MF-Hybrid network and GLGAN+VIT achieve two optimal scores in restoring the original-sized murals, the practical image restoration results unmistakably favor the MF-Hybrid network, highlighting its superior performance.

\subsubsection{Refinement of the giant mural}
After completing the restoration of the regular size mural cuttings, the results of the ablation experiments on the giant mural are shown in Fig.7.

Directly stitching small images together, especially for large areas of damage, results in noticeable seams in the stitched giant mural shown in Fig.7.(c). However, by employing a multi-perspective strategy and averaging and fusing 16 different versions, the stitch marks are effectively smoothed out shown in Fig.7.(d). This approach has demonstrated particularly good restoration results for large damaged areas.

The multi-scale fusion method combines the advantages of the cropping and downsizing methods to utilize the restoration results at different scales. As depicted in the results shown in Fig.7.(e), the cropping method effectively preserves the detailed information of the restoration results while also maintaining the structure of the original image. Therefore, the multi-scale completion strategy has proven to be highly effective for the restoration of large murals, particularly when dealing with extensive areas of damage.

\subsection{Robustness analysis}

When assessing the robustness of the 3M algorithm, due to the limited quantity of Yongle Palace murals, which cannot be replaced, we augmented the mask coverage to test the model's robustness. The results are presented in TABLE \MakeUppercase{\romannumeral 2}, where the mask sizes correspond to mural coverage areas of 37.33\%, 46.05\%, and 57.16\%, respectively. It can be observed that despite the increasing mask coverage, the restoration quality did not significantly decline, indicating the stability of the proposed 3M-Hybrid model.

\begin{table}[]
    \caption{Experimental results on the robustness of the model with different defect coverage rates\label{tab:table1}}
    
    \begin{tabular}{ccccc}
    \toprule[2pt]
    \makebox[0.07\textwidth][c]{MASK SIZE} & \makebox[0.07\textwidth][c]{PSNR} & \makebox[0.07\textwidth][c]{SSIM} & \makebox[0.07\textwidth][c]{MSE} & \makebox[0.07\textwidth][c]{MAE} \\ \hline
    
    \hline
              
    37.33\% & 22.9146 & 0.7396 & 332.3703 & 11.8511 \\
    46.05\% & 21.6699 & 0.7159 & 442.6822 & 13.8805 \\
    57.72\% & 20.6072 & 0.6810 & 565.4011 & 18.9041 \\ \hline
    
    \bottomrule[2pt]
    \end{tabular}
\end{table}

\section{Discussion and conclusion}

\subsection{conclusion:}

This study presents the 3M-Hybrid model for the restoration of the giant wall paintings at the Yongle Palace. The term "3M" refers to three key strategies: multi-frequency, multi-perspective and multi-scale, and "Hybrid" refers to the hybrid CNN-VIT network. 

Among them, the multi-frequency strategy and hybrid CNN-VIT network constitute a regular-sized mural restoration network named MF-Hybrid network for multiple types and scales of imperfections in small datasets. The hybrid CNN-VIT network optimises the network structure and enhances feature extraction for small-scale murals. The multi-frequency strategy optimises training data effectively exposes features in different frequency bands, enabling accurate and efficient learning from the limited small dataset. 

In giant mural assemblage, this study proposes a multi-perspective and multi-scale strategy. The multi-perspective strategy addresses stitching gaps that may arise when combining smaller images, ensuring a seamless view for mural restoration. The multi-scale strategy enhances the model's ability to learn the overall image structure and effectively repair oversized defects.

Through the integration of these four strategies, the 3M-Hybrid model demonstrates remarkable restoration performance for giant murals. It not only provides effective restoration methodologies but also offers valuable insights for future restoration efforts on large-scale mural artworks. The 3M-Hybrid model stands as a significant contribution in the field of giant mural restoration.

\subsection{limitations:} 

First, the proposed approach relies on conducting multiple experiments to select the optimal values for three scale fusion weights. However, this method may not be precise enough, considering that the weight settings encompass countless possibilities, while the number of experiments is limited. Consequently, the weight values determined based on experimental outcomes can only guarantee relatively favorable final results.

Second, the evaluation metrics used are not sufficiently objective. The current four evaluation metrics lack a comprehensive assessment of the image structure, often failing to accurately reflect human perception and evaluation of the images. This limitation is also why the evaluation of many image restoration results, especially in the field of art restoration, requires expert assessment rather than relying solely on mathematical calculations using metrics.

\section{Appendix}

The following four tables show the restoration results obtained by the CE model, GLGAN model, GMCNN model, and PEN model, respectively, after applying four masks to the Yongle Palace murals. Based on the results presented in the four tables, the GLGAN model performs the best in restoring normal-sized murals in the Yongle Palace. Therefore, the GLGAN model has been selected as the fundamental CNN structure for the proposed 3M-Hybrid Model in this research.

\begin{table}[!ht]
    \setlength\tabcolsep{3pt}
    \centering
    \caption{Evaluation metrics comparison of free shape block masks}
   
    \begin{tabular}{ccccccc}
    \toprule[2pt]
    \makebox[0.02\textwidth][c]{} &\makebox[0.02\textwidth][c]{Model} &\makebox[0.03\textwidth][c]{10\%} & \makebox[0.03\textwidth][c]{15\%} & \makebox[0.03\textwidth][c]{20\%} & \makebox[0.03\textwidth][c]{25\%} & \makebox[0.03\textwidth][c]{30\%} \\ \hline
    
    \hline
        
        ~ & CE & 0.029855 & 0.052917 & 0.065413 & 0.084941 & 0.098137  \\ 
        MAE & GLGAN & \pmb{0.025384} & \pmb{0.043237} & \pmb{0.056224} & \pmb{0.072528} & \pmb{0.088556}  \\ 
        ~ & GMCNN & 0.027092 & 0.046117 & 0.059091 & 0.075859 & 0.092188  \\ 
        ~ & PEN & 0.029498 & 0.050581 & 0.066059 & 0.083171 & 0.103418  \\ \hline
        ~ & CE & 0.017069 & 0.03105 & 0.037069 & 0.048446 & 0.055397  \\ 
        MSE & GLGAN & \pmb{0.013302} & \pmb{0.023217} & \pmb{0.030382} & \pmb{0.039244} & \pmb{0.048059}  \\ 
        ~ & GMCNN & 0.015406 & 0.026612 & 0.034215 & 0.043604 & 0.053318  \\ 
        ~ & PEN & 0.01772 & 0.030228 & 0.040848 & 0.050037 & 0.064328  \\ \hline
        ~ & CE & 18.762274 & 15.782112 & 15.291915 & 13.793731 & 13.330272  \\ 
        PSNR & GLGAN & \pmb{19.98415} & \pmb{17.439302} & \pmb{16.329344} & \pmb{14.934479} & \pmb{14.076589}  \\ 
        ~ & GMCNN & 19.233339 & 16.776512 & 15.761546 & 14.423636 & 13.595858  \\ 
        ~ & PEN & 18.685036 & 16.211851 & 15.041623 & 13.869326 & 12.617146  \\ \hline
        ~ & CE & 0.891139 & 0.818218 & 0.771656 & 0.70763 & 0.657839  \\ 
        SSIM & GLGAN & \pmb{0.906947} & \pmb{0.848489} & \pmb{0.797811} & \pmb{0.746036} & \pmb{0.689936}  \\ 
        ~ & GMCNN & 0.903058 & 0.842724 & 0.790927 & 0.738731 & 0.682054  \\ 
        ~ & PEN & 0.89661 & 0.831848 & 0.779768 & 0.7238 & 0.662579  \\ \hline
    \end{tabular}
\end{table}

\begin{table}[!ht]
    \setlength\tabcolsep{3pt}
    \centering
    \caption{Evaluation metrics comparison of free jelly-like masks}
    
    \begin{tabular}{ccccccc}
    \toprule[2pt]
    \makebox[0.02\textwidth][c]{} &\makebox[0.02\textwidth][c]{Model} &\makebox[0.03\textwidth][c]{20\%-30\%} & \makebox[0.03\textwidth][c]{30\%-40\%} & \makebox[0.03\textwidth][c]{40\%-50\%} & \makebox[0.03\textwidth][c]{50\%-60\%} & \makebox[0.03\textwidth][c]{60\%-70\%} \\ \hline
    
    \hline
        ~ & CE & 0.072578 & 0.119536 & 0.148029 & 0.190882 & 0.211071  \\ 
        MAE & GLGAN & \pmb{0.044712} & \pmb{0.078114} & \pmb{0.082603} & \pmb{0.113803} & \pmb{0.15444}  \\ 
        ~ & GMCNN & 0.051245 & 0.088477 & 0.094394 & 0.129584 & 0.173384  \\ 
        ~ & PEN & 0.076389 & 0.130631 & 0.139712 & 0.189978 & 0.235491  \\  \hline
        ~ & CE & 0.03807 & 0.063202 & 0.08263 & 0.106168 & 0.113475  \\ 
        MSE & GLGAN & \pmb{0.017758} & \pmb{0.032716} & \pmb{0.033232} & \pmb{0.048487} & \pmb{0.072096}  \\ 
        ~ & GMCNN & 0.022512 & 0.041234 & 0.042214 & 0.061021 & 0.088158  \\ 
        ~ & PEN & 0.045219 & 0.077972 & 0.079593 & 0.112083 & 0.141258  \\  \hline
        ~ & CE & 14.890925 & 12.644198 & 11.348081 & 10.308614 & 10.153835  \\ 
        PSNR & GLGAN & \pmb{18.423899} & \pmb{15.764747} & \pmb{15.600578} & \pmb{14.000488} & \pmb{12.237443}  \\ 
        ~ & GMCNN & 17.149782 & 14.553966 & 14.346415 & 12.79388 & 11.178774  \\ 
        ~ & PEN & 14.346288 & 11.913223 & 11.76566 & 10.294182 & 9.205093  \\  \hline
        ~ & CE & 0.628776 & 0.473309 & 0.312482 & 0.240706 & 0.218685  \\ 
        SSIM & GLGAN & \pmb{0.799595} & \pmb{0.678711} & \pmb{0.664797} & \pmb{0.576311} & \pmb{0.430019}  \\ 
        ~ & GMCNN & 0.772676 & 0.64272 & 0.619654 & 0.526178 & 0.379073  \\ 
        ~ & PEN & 0.67176 & 0.512505 & 0.46465 & 0.365215 & 0.26429  \\ \hline
    \end{tabular}
\end{table}

\begin{table}[!ht]
    \setlength\tabcolsep{3pt}
    \centering
    \caption{Evaluation metrics comparison of free jelly-like masks}
  
    \begin{tabular}{ccccccc}
    \toprule[2pt]
    \makebox[0.02\textwidth][c]{} &\makebox[0.02\textwidth][c]{Model} &\makebox[0.03\textwidth][c]{10\%-20\%} & \makebox[0.03\textwidth][c]{20\%-30\%} & \makebox[0.03\textwidth][c]{30\%-40\%} & \makebox[0.03\textwidth][c]{40\%-50\%} & \makebox[0.03\textwidth][c]{50\%-60\%} \\ \hline
    
    \hline
        ~ & CE & 0.045187 & 0.078194 & 0.106997 & 0.137725 & 0.179821  \\ 
        MAE & GLGAN & \pmb{0.034987} & \pmb{0.059347} & \pmb{0.083107} & \pmb{0.110185} & \pmb{0.143224}  \\ 
        ~ & GMCNN & 0.039175 & 0.066457 & 0.092122 & 0.121857 & 0.156982  \\ 
        ~ & PEN & 0.049206 & 0.081598 & 0.114519 & 0.152526 & 0.197999  \\  \hline
        ~ & CE & 0.024391 & 0.042449 & 0.056904 & 0.075231 & 0.096913  \\ 
        MSE & GLGAN & \pmb{0.016853} & \pmb{0.028444} & \pmb{0.039723} & \pmb{0.054991} & \pmb{0.070656}  \\ 
        ~ & GMCNN & 0.021079 & 0.035424 & 0.04843 & 0.066556 & 0.084114  \\ 
        ~ & PEN & 0.029981 & 0.048779 & 0.067874 & 0.092933 & 0.116352  \\  \hline
        ~ & CE & 17.272425 & 14.323305 & 13.109604 & 11.942086 & 10.817717  \\ 
        PSNR & GLGAN & \pmb{18.974913} & \pmb{16.222977} & \pmb{14.793048} & \pmb{13.410428} & \pmb{12.366553}  \\ 
        ~ & GMCNN & 17.895704 & 15.227664 & 13.873782 & 12.519254 & 11.497271  \\ 
        ~ & PEN & 16.507465 & 13.870496 & 12.488193 & 11.070246 & 10.124557  \\  \hline
        ~ & CE & 0.820638 & 0.691934 & 0.596612 & 0.500815 & 0.351323  \\ 
        SSIM & GLGAN & \pmb{0.863249} & \pmb{0.761517} & \pmb{0.684189} & \pmb{0.605511} & \pmb{0.471572}  \\ 
        ~ & GMCNN & 0.852506 & 0.742125 & 0.658225 & 0.577012 & 0.434665  \\ 
        ~ & PEN & 0.828192 & 0.698484 & 0.604602 & 0.504259 & 0.339922  \\ \hline
    \end{tabular}
\end{table}

\begin{table}[!ht]
    \centering
    \caption{Evaluation metrics comparison of free linear masks}
   
    \begin{tabular}{ccccc}
    \hline
        \textbf{MODEL} & \textbf{MAE} & \textbf{MSE} & \textbf{PSNR} & \textbf{SSIM } \\ \hline
        CE & 0.057742 & 0.068737 & 12.058658 & 0.507938  \\ 
        GLGAN & 0.039058 & 0.039252 & \pmb{14.778023} & 0.747997  \\ 
        GMCNN & \pmb{0.037734} & \pmb{0.038402} & 14.659587 & \pmb{0.752893}  \\ 
        PEN & 0.042668 & 0.046873 & 14.019386 & 0.728164  \\ \hline
    \end{tabular}
\end{table}

\newpage
\bibliographystyle{IEEEtran}
\small\bibliography{reference}

%

\end{document}